\documentclass[nonacm, acmtog]{acmart}

\usepackage{booktabs} % For formal tables
\usepackage[capitalize,nameinlink]{cleveref}
\usepackage{soul}
\usepackage{xfrac}
\usepackage{xcolor}
\usepackage{caption}
\usepackage{paralist}
\usepackage{multirow}

\usepackage[ruled]{algorithm2e} % For algorithms

\SetAlFnt{\small}
\SetAlCapFnt{\small}
\SetAlCapNameFnt{\small}
\SetAlCapHSkip{0pt}

\DeclareGraphicsExtensions{.png,.jpg,.pdf,.ai,.psd}
\def\etal{{et al.}}
\def\ie{{i.e.}}
\def\eg{{e.g.}}

\title{Overpainting: Localized Context-aware Diffusion Image Editing}

\author{Sam Sartor}
\affiliation{%
  \institution{College of William \& Mary}
  \city{Williamsburg}
  \country{USA}
}
\orcid{0009-0001-1915-6887}
\email{ssartor@adobe.com}

\author{Iliyan Georgiev}
\affiliation{%
    \institution{Adobe Research}
    \country{United Kingdom}
}
\email{igeorgiev@adobe.com}
\orcid{0000-0002-9655-2138}

\author{Michael Fischer}
\affiliation{%
    \institution{Adobe Research}
    \country{United Kingdom}
}
\email{mifischer@adobe.com}
\orcid{0000-0002-2610-4831}

\author{Valentin Deschaintre}
\affiliation{%
    \institution{Adobe Research}
    \country{United Kingdom}
}
\email{deschain@adobe.com}
\orcid{0000-0002-6219-3747}

\author{Pieter Peers}
\affiliation{%
  \institution{College of William \& Mary}
  \city{Williamsburg}
  \country{USA}
}
\orcid{0000-0001-7621-9808}
\email{ppeers@siggraph.org}

\newcommand\mtx[1]{\mathbf{#1}}
\newcommand\vect[1]{\mathbf{#1}}

\def\W{\mtx{W}}
\def\A{\mtx{A}}
\def\B{\mtx{B}}

\def\R{\mathbb{R}}
\def\X{\vect{x}}
\def\H{\vect{y}}

\def\Attn{\mathtt{Attn}}

\begin{document}

%%%%%%%%%%%%%%%%%%%%%%%%%%%%%%%%%%%%%%%%%%%%%%%%%%%%%%%%%%%%
% Teaser figure
%%%%%%%%%%%%%%%%%%%%%%%%%%%%%%%%%%%%%%%%%%%%%%%%%%%%%%%%%%%%
\begin{teaserfigure}
    \centering
    \includegraphics[width=\textwidth]{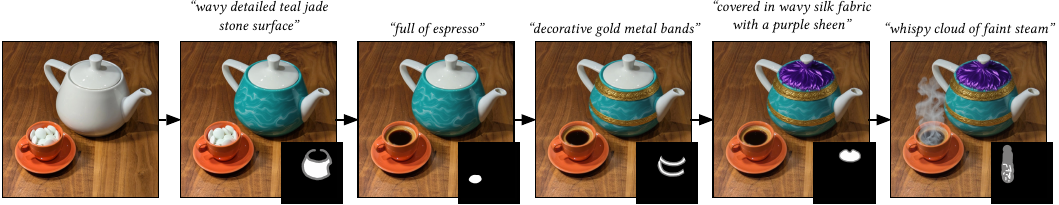}
    \vspace{-0.3cm}
    \caption{A sequence of localized overpainting edits. We propose to use a trimap as input, with gray representing areas where change \emph{may} occur and white where change \emph{must} occur.
    This provides strong edit localization while simplifying the masking task, as it does not need to be pixel accurate thanks to the gray regions handling borders.}
    \label{fig:teaser}%
\end{teaserfigure}

%%%%%%%%%%%%%%%%%%%%%%%%%%%%%%%%%%%%%%%%%%%%%%%%%%%%%%%%%%%%
% Abstract and
% Publication categories & keywords
%%%%%%%%%%%%%%%%%%%%%%%%%%%%%%%%%%%%%%%%%%%%%%%%%%%%%%%%%%%%
%%%%%%%%%%%%%%%%%%%%%%%%%%%%%%%%%%%%%%%%%%%%%%%%%%%%%%%%

\begin{abstract}
  We present ``overpainting'', an image editing operation which
  offers both control over the location of the edit
  %(like inpainting) => PP dangerous; this will result in inpainting reviewers
  and awareness of the previous content in
  that location. %(like instruction-guided editing).  => PP same reason
  The overpainted
  area is given by a trimap, where white-annotated pixels \emph{must} be
  edited, gray-annotated pixels \emph{may} be edited, and black-annotated
  pixels \emph{must not} be edited. This enables both precise and
  loose control, depending on user intent.
  
  We implement overpainting by adapting a pretrained image editing diffusion model
  using a combination of joint attention and low-rank adaption across
  input images %%(cf. Teamwork~\cite{Sartor:2025:TCD})  %% PP: confusing whether it refers to the combination or the LoRA across images.  Maybe better for the main text than the abstract
  with
  attention-dropout to balance the information flow between noise,
  source and mask images. We present a novel, automated,
  training data generation pipeline that (1)~generates a set of
  candidate image pairs leveraging existing language-based editing
  models, (2)~carefully curates those pairs, and (3)~extracts a
  trimap from each usable pair.  We demonstrate the versatility of our
  overpainting model on a wide range of editing tasks.
\end{abstract}
\maketitle

%%%%%%%%%%%%%%%%%%%%%%%%%%%%%%%%%%%%%%%%%%%%%%%%%%%%%%%%%%%%
\section{Introduction}
%%%%%%%%%%%%%%%%%%%%%%%%%%%%%%%%%%%%%%%%%%%%%%%%%%%%%%%%%%%%

Image diffusion models allow users to create fantastic images by
simply providing a textual description of the picture in their mind.  As
language is an imprecise tool to describe fine details, diffusion
models necessarily must hallucinate that missing information.  This
generative behavior makes diffusion models an ideal
foundation for editing tools. %
Recently, dedicated image editing diffusion
models~\cite{BFL:2025:FKF,Zimage:2025:ZIA,Esser:2024:SRF,Wu:2025:QIT}
have emerged, allowing the application of high-level
identity-preserving editing operations based on multi-modal input such
as a source image and a prompt describing the change. However,
obtaining satisfactory edits with precise user-directed localization,
such as those shown in~\autoref{fig:teaser}, requires careful
prompt engineering. Even for multi-modal models provided with a 
mask, the models see the mask more as a suggestion than a
constraint.

\begin{figure}
    \centering
    \includegraphics[width=\columnwidth]{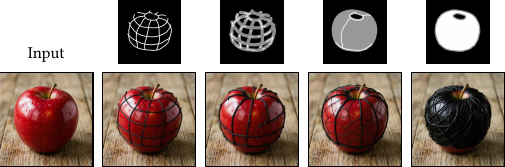}
    %\vspace{-3mm}
    \caption{Trimap variations on an image of an
      apple with the prompt \emph{``wrapped in black twine''}. 2nd
      column: the apple is wrapped by a carefully drawn mask without
      any gray regions, and the prompt will be exactly applied in the
      masked regions only. 3rd column: adding gray regions around
      the mask allows overpainting to make local changes to improve
      the quality of the results (\ie, more variations in the
      silhouette of the thread). 4th column: marking a large region
      of the apple with a gray mask and only including a few white
      lines provides much greater freedom for overpainting, and
      consequently less control on the actual content by the user.
      Last column: marking most of the apple white results in a very
      densely wrapped apple.}
    \label{fig:maskvariations}%
\end{figure}

In this paper we introduce a novel diffusion-based editing operation
\emph{overpainting}, which makes local alterations to an image while
respecting the previous content (\eg, partially altering an object's
surface to be made of stone, or to be covered in hair, as shown in \cref{fig:teaser,fig:maskvariations}). After overpainting, the viewpoint,
composition, and style of an image will be the same, but depicted
objects or parts of objects must differ. Each individual pixel is
classified as \emph{edited} if its color or other features have been
meaningfully altered, and \emph{unedited} otherwise. The key challenge
of overpainting is control: allowing the user to specify both which
pixels should be edited (via the mask) and how they should be edited
(via the prompt).

%
%%A further challenge is usability.   %%% PP: this is not a good way to start a paragraph, especially given the previous one where you did not introduce a challenge but the method.  Better to drop this sentence.
Specifying an exact editing mask
around existing image features is cumbersome and error-prone (\eg, for
thin features or precise object boundaries). It is also often
ambiguous for edits with ill-defined boundaries (\eg, frost forming on
a window or smoke dissipating) and for to-be-created features whose
location depends on the target edit itself (\eg, whiskers protruding
from an object on which the user asked the model to grow hair). We
tackle this challenge by introducing a trimap that marks not only pixels
which \emph{must} be edited (white) and must not be edited (black), but also
pixels that \emph{may or may not} be edited (gray; \cref{fig:maskvariations}).
% This approach balances the need for accurate masks versus the user's ability to provide fine-grained positional control.  
%\MF{Many overpainting edits have no well-defined boundary: frost forming on a window, moss thinning on a stone, or smoke dissipating. 
%The trimap's gray region addresses this by letting the user mark a permitted neighborhood within which the model itself decides where the edit ends and how it blends out in a semantically meaningful way — a level of delegation that a binary mask cannot express.}
In contrast to an inexact binary mask where the marked region has an ambiguous
meaning (\ie, it combines \emph{``must''} and \emph{``may be''} changed regions),
a trimap provides unambiguous semantic meaning for editing each region, allowing the user to apply different strategies to localize edits: binary masks for providing exact control, rough masks
with wide gray margins for 'loose' control, or strategically placed thin gray masks around intricate details to provide the editing model freedom to blend or modify the intricate details naturally.

% \MF{in this paragraph, we explain how we enable overpainting -- via the trimap --, and in the subsequent paragraph we explain the technical realization (teamwork, flux). I would put more emphasis on why we NEED overpainting in the first place, and give more examples or a stand-alone definition} -- added the above sentence, to clarify 

We base our overpainting model on a pretrained editing diffusion
model~\cite{BFL:2025:FKF}, and adapt the model to adhere to both the trimap and the image content
while retaining the embedded priors as much as possible.  To achieve this goal, we balance different information sharing strategies between the different input
images (source, mask, and noise, each representing an
``image-stream'').  In particular, we combine joint attention and
Teamwork (\ie, low-rank adaptors (LoRA) \emph{across} input
images~\cite{Sartor:2025:TCD}).  We observe that both information
sharing strategies trade-off complementary aspects: editing quality
versus mask adherence.  Therefore, we combine joint attention with
Teamwork and leverage attention-dropout to balance the flow of
information sharing across both mechanisms.

Existing image editing datasets (\eg, MagicBrush~\cite{Zhang:2023:MMA}
and Pico-Banana-400K~\cite{Qian:2025:PB4}) are dominated by
non-over\-paint\-ing editing operations such as object insertion,
removal, and global edits, making such datasets unsuited for training
our overpainting model. %\MF{Here the reader will ask: why are they unsuited, what even is overpainting?}
Instead, we leverage existing editing models to
generate a diverse set of paired training exemplars.  Such models are
prone to a number of failures (\ie, under-editing, over-editing, and
shifting) for which we propose an automatic detection and filtering approach. Finally, we
introduce a high-quality hand-crafted overpainting validation set of
$180$ exemplars.

We demonstrate the versatility of our overpainting model on a wide
range of editing examples that are not possible with existing image
editing diffusion models. \\
In summary our contributions are:
\begin{compactenum}
\item overpainting: a novel image editing task utilizing trimaps
which combines local control with content-awareness;
\item a combined information sharing strategy between image-streams
  for balancing mask adherence versus edit quality control; and
\item an automated pipeline for generating overpainting training
  pairs, and a hand-crafted overpainting benchmark comprising $180$
  exemplars.
\end{compactenum}

\section{Related Work}
\label{sec:related}

\paragraph{Diffusion-based Image Editing}
The diversity and quality of images that can be generated with
diffusion denoising models~\cite{Ho:2020:DDP,Rombach:2022:HRI},
especially when conditioned on prompt embeddings, has caught the
attention of users and researchers alike. Recent advances have
improved controllability~\cite{Zhang:2023:ACC} and have seen a gradual
shift to more powerful transformer models~\cite{Peebles:2023:SDM}.
Whereas the original image diffusion models were focused on modeling
the data distribution of natural images (\ie, creation), a new class
of image \emph{editing} models~\cite{Avrahami:2025:SFV,
  Brooks:2023:IP2P, Cao:2023:MDS, Hertz:2022:PTP, Kawar:2023:IMA,
  Lin:2024:TDI, Parmar:2023:PPZ} focuses on modeling conditional
transformations. Editing models take as input a source image alongside
a prompt describing the desired change, and output a modified image
that reflects that change.  Recent models have shown great progress in
prompt adherence~\cite{BFL:2025:FKF},
efficiency~\cite{Zimage:2025:ZIA}, flexibility and
expandability~\cite{Esser:2024:SRF}, and text
rendering~\cite{Wu:2025:QIT}.  We refer the reader to the excellent
survey by Huang~\etal~\shortcite{Huang:2025:DMB} for a more in-depth
overview.  While image-editing models are good at making global edits,
it is hard to make localized changes without global side-effects (\eg,
image and colors shifts, changes outside the target area, etc.)  In
this paper, we build on existing image-editing foundation models and
introduce a new type of localized editing operation, namely
\emph{overpainting}.

\paragraph{Inpainting}
As its name suggests, our overpainting paradigm is somewhat related to
inpainting, which aims to seamlessly (re)syn\-the\-size a masked region in
an image, conditioned on the content of the surrounding area and a
prompt~\cite{Manukyan:2025:HPH,Rombach:2022:HRI,Yang:2023:UAU,Zhuang:2024:TWO,Lugmayr:2022:REI,Podell:2024:SDXL,Saharia:2022:PII,Liu:2021:PDG,Wang:2025:TEI}. Inpainting
is intended to be used for localized editing, but ignores the
existing, masked content, regenerating it from scratch. This
necessarily loses the identity of what was masked and may produce unintended ``side-edits'' in the masked region. Our
overpainting addresses this shortcoming by allowing the
generation to be both precisely located and directly conditioned on the existing context from the masked region.

\paragraph{Localized Editing}
Blended diffusion~\cite{Avrahami:2022:BDT,Avrahami:2023:BLD}
introduces local control by blending a noised version of the original
image with CLIP-guided latents. However, blended diffusion suffers
from latent drift (\eg, loosing details in the unmasked regions) and
boundary inconsistencies.  BrushEdit~\cite{Li:2024:BAI} unifies
inpainting and editing via an agent-based, free-form, interactive
framework. Both blended diffusion and BrushEdit suffer from boundary bleeding in case of inaccurately specified masks.
Layered diffusion brushes~\cite{Gholami:2025:SIE} enable interactive
mask and prompt-based editing.  A key innovation is that the method
generates distinct layers with transparency per edit operation.  While
the quality of the results is less sensitive to the accuracy of the
mask, an accurate mask is still required to ensure no unwanted edits
outside the intended target.  To alleviate the dependency on mask
accuracy, several methods infer a region of interest from the prompt.
\citet{Couairon:2023:DDS} use a contrastive noise process to
internally infer masks. \citet{Cvejic:2025:PFG} learn specialized
tokens based on segmentation to localize edits.  \citet{Hu:2025:IEA}
determine localization via chain-of-thought reasoning and VLMs.
\citet{Simsar:2025:LIE} leverage intrinsic semantic encoding of
intermediate layers to define a region of interest (ROI) that serves
as boundary for attention regularization (\ie, penalizing unrelated
tokens inside the ROI).  While powerful, these mask-free methods are
inherently limited to recognizable parts, and finer-grained edits are
challenging.  
%We also recognize the difficulty of providing exact masks. 
To balance mask accuracy and precise localization, we take
inspiration from image
matting~\cite{Yu:2021:HRD,Xu:2017:DIM,Chuang:2001:BAD,Sun:2004:PM} and
allow the user to specify the ROI via a trimap.

\paragraph{Texture and Material Painting}
A specialized set of image editing operations focuses on changing the
texture or material of an object or
region. CoatFusion~\cite{Levy:2025:CFC} allows to 'coat' an object
with a new material. Unlike material transfer, CoatFusion keeps the
fine-details of the underlying object.  Texture
painting~\cite{Ardelean:2025:EFP,Hu:2024:DTP,Pandey:2025:P3G,Huang:2025:REO,Chen:2024:ZSI}
allows the user to apply a texture from an exemplar to a ROI in a
target image while following the underlying geometry.
MatSwap~\cite{Lopes:2025:MSL} and
MaterialFusion~\cite{Garifullin:2025:MFH} further expand on the idea
of texture painting by adding light and material awareness to the
transfer. All these methods can be seen as specializations of
overpainting that focus on a particular task (\eg, material coating or
texture painting).  UltraEdit~\cite{Zhao:2024:UIB} and
AlterBute~\cite{Reiss:2026:AEI} support changing intrinsic properties,
such as material, texture and even shape of a marked object given a
prompt and a foreground and background mask. For both UltraEdit and
Alterbute, inaccuracies in the foreground mask lead to boundary
bleeding.  Overpainting departs from the notion of a strict foreground
and background mask, and instead leverages a trimap that marks the
user's (un)certainty in the application of the editing
operation in a region.

\section{Method}
\label{sec:method}

The input to overpainting is a source image, a noisy latent,
a prompt describing the edit, and a trimap that marks image regions
that either must be altered to match the prompt (white),
may be edited (gray), or that should not be modified (black).
The output is the edited image. %
We show that existing pre-trained editing models can be adapted to
overpainting, without requiring complete re-training. In particular,
we adapt the FLUX.1 Kontext model ~\cite{BFL:2025:FKF}.

\paragraph{LoRA Adapted Joint Attention}
FLUX.1 Kontext already supports multiple image-streams via a 3D
positional encoding (in which the third dimension is the image
index). Information is exchanged between streams via joint attention:
$\Attn(\H_{prompt}, \H_{target}, \H_{source})$, where $\H_{target}$
and $\H_{source}$ are the features for the existing input and edited images.
A third stream of image tokens $\H_{mask}$ can be handled the same way.
Then, to adapt the semantic interpretation of the three
image-streams, a common strategy would be to apply low-rank adaption (LoRA)
\cite{Hu:2022:LRA} on the linear layers.  Formally, given the frozen
linear weights $\W \in \R^{m \times n}$ of a layer, the adapted
weights are expressed as: $\W + \Delta \W$, with $\Delta \W = \A \B$,
$A \in \R^{m \times r}$, and $\B \in \R^{r \times n}$ forming a low
rank approximation, and thus: $\H_s = \W \X_s + \A\B \X_s$, with
$s \in \{target, source, mask\}$.

\paragraph{Teamwork} An alternative strategy for exchanging
information between different image-streams is
Teamwork~\cite{Sartor:2025:TCD} which also applies low-rank
adaptation on each linear layer of the model, but instead of a single LoRA for all image-streams,
computes a joint-LoRA \emph{across} the different
image-streams (called teammates), followed by a
separate per-stream self-attention: $\Attn(\H_s)$.  Hence, information
is not exchanged via joint attention, but through the linear
layers. Formally, given the inputs $\X_s$, Teamwork computes the
outputs $\H_s$ for each $s \in \{target, source, mask\}$ as:
$\H_s = \W \X_s + \B_s (\A_{target} \X_{target} + \A_{source}
\X_{source} + \A_{mask} \X_{mask})$\footnote{Sartor and
  Peers~\shortcite{Sartor:2025:TCD} employ the equivalent notation:
  $\H = \mathtt{block}(\W) + \B \A \X$, where $\H, \X, \A$, and $\B$
  are the stacked tensors of the corresponding tensors from the
  target, source, and mask streams, \eg,
  $\H = [\H_{target} | \H_{source} | \H_{mask}]$.}.

\paragraph{Combining Joint Attention and Teamwork}
Empirically we find that LoRA adapted joint attention and Teamwork
provide complementary benefits.  A LoRA adapted joint attention model
produces high quality edits, but it does not adhere well to the
user-specified mask (\autoref{tab:overpainting}, \emph{'FLUX+LoRA'}).
Conversely, a Teamwork-only model adheres well to the mask, but
provides lower edit quality (\autoref{tab:overpainting},
\emph{'Teamwork'}).  Intuitively, Teamwork imposes a strong positional
bias (\ie, only latents with the same (x,y) position exchange
information) which benefits mask adherence. In contrast, LoRA adapted
joint attention can attend between pixels at different positions in
the different image-streams, providing greater flexibility in applying
the requested edit operation. The two approach coordination from opposite
directions: FLUX+LoRA communicates through the attention layers, Teamwork
through the linear layers. This raises the question if combining both (\ie,
using Teamwork's LoRA adaptation across image-streams and  also joint
attention across image-streams) can combine their advantages.
However, we found that na\"ively combining both strategies does not
improve performance (\autoref{tab:overpainting} \emph{'naive'}). We
posit that since training starts from FLUX.1 Kontext which already has
converged joint attention, there is a bias towards further optimizing
this information sharing channel. To better balance the fine-tuning of both
information sharing channels, we leverage attention-masking to randomly dropout the
attention from one stream to another, leaving Teamwork as the only mechanism for
coordination some percentage of the time. Empirically, we determined that a $50\%$ attention-dropout
rate strikes a good balance between edit quality and mask adherence (see also
\autoref{sec:eval}).

\begin{figure*}[t]
    \centering
    \includegraphics[width=.99\textwidth]{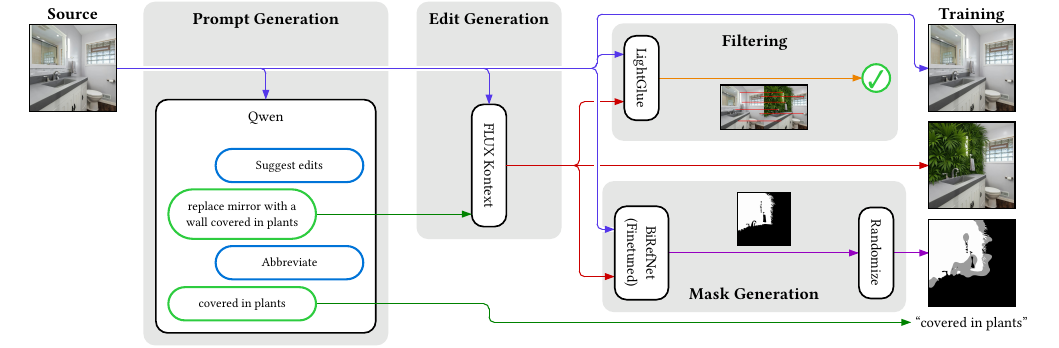}
    \vspace{-3mm}
    \caption{ Overview of the training data generation pipeline. Given
      an image sourced from the public photo collection website
      Pexels, an editing prompt is generated using Qwen. This prompt,
      together with the source image, is subsequently fed to a
      pretrained image editing diffusion model (FLUX.1 Kontext), and
      automatically checked for common failure modes such as
      over-editing, under-editing and shifting.  If approved, the
      source and edited image are passed into a mask estimation
      network.  Finally, during training, we convert the estimated
      alpha mask into a trimap.}
    \label{fig:pipeline}%
\end{figure*}

\paragraph{Automated Training Data Generation}
To train our overpainting model we require paired data in the form of
an original image, an edited version, a mask, and a prompt describing
the edit operation.  We follow the well-established process of
leveraging VLMs and image diffusion models to
generate training data~\cite{Brooks:2023:ILF,
  Zhang:2023:MMA,Wasserman:2025:PBI,Qian:2025:PB4}.

We randomly sample $11,\!323$ freely licensed photographs from
\url{www.pexels.com} as source photographs. Each image is then passed
to the Qwen2.5-VL-72B multimodal language model~\cite{Wu:2025:QIT}
which is asked to suggest possible editing prompts.

Next, we feed the generated prompts and the corresponding source image
into FLUX.1 Kontext to produce the edited photographs. While powerful,
FLUX.1 Kontext is not perfectly reliable and its outputs exhibit
several common failure modes.  We found that on average $6.4\%$ of the
generated results are nearly identical to the original image (\ie,
under-editing). In $7.1\%$ of the cases, FLUX.1 Kontext ignores the
source image and generates an entirely new image or makes drastic
stylistic global changes without positional locality (\ie,
over-editing).  While the remainder of the results feature a suitable
edit, $42.7\%$ exhibit an additional global non-affine shift of the
whole image. %
We filter out over- and under-edited results by finding the most and
least changed $16\times16$ image patch.  For the most-changed patch we
require a minimum $75\%$ ratio of change and for the least-changed
less than $15\%$ should be changed; we reject the candidate exemplar
if these conditions are not met.  To filter misaligned images, we
extract matching key points using
LightGlue~\cite{Lindenberger:2023:LGL}, and reject pairs where more
than $15\%$ of the key points have moved more than $4$ pixels or too
few ($<5$) matching key points are found.

\begin{figure}
    \centering
    \includegraphics[width=\linewidth]{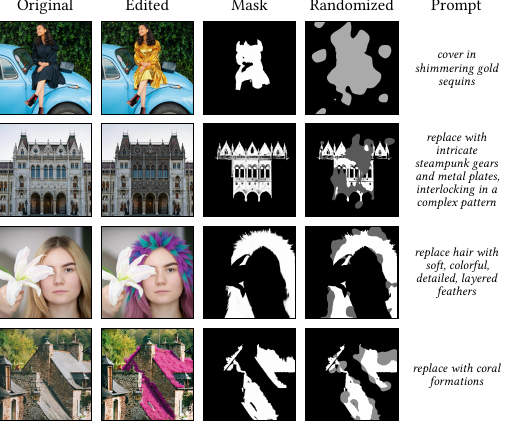}
    \vspace{-6mm}
    \caption{Examples of automatically generated paired training
      images with corresponding prompts and on-the-fly generated randomized trimaps.}
    \label{fig:training}%
\end{figure}

To generate corresponding trimap masks, we adapt BiRefNet~\cite{Zheng:2024:BRH}
to accept both the original and edited images, and finetune it to
precisely segment the edit. We found that BiRefNet's priors on
foreground/background separation translate naturally.
\cref{fig:training} shows a selection of estimated binary masks and
corresponding image pairs. Our finetuned BiRefNet accurately distinguishes
meaningful edits from noise, even on fine details like hair.
We further degrade those precise masks to
mimic user-drawn trimaps. We do so on-the-fly, during training, by
creating randomized gray ``may edit'' regions of random width and
center around the boundary of the binary mask.

\cref{fig:pipeline} summarizes the automated data generation
process. We refer to the appendices for the implementation
details.

%\paragraph{Training}
%
%To train our overpainting model, we use the standard diffusion loss
%and the paired training images with corresponding prompts and
%on-the-fly generated trimaps.  We train our model for $8,\!000$ steps
%using the standard settings of the %Prodigy optimizer
%\cite{Mishchenko:2023:PEA}.

\paragraph{Secondary Task: inpainting}
When testing the utility of overpainting with a control group, we
observe that artists like to combine both overpainting and inpainting
operations to achieve their vision.  To cater to both editing
paradigms, we also train our overpainting model on the secondary task
of inpainting by randomly inserting a inpainting training exemplar
with $15\%$ probability. Each inpainting exemplar consists of a random
Pexels image as target image, and as input the same image with a
random shape of area zeroed-out. %%%%\PP{The zeroed-out is very important!!!!}
We use the prompt ``inpaint'' or ``inpaint. \{VLM description\}''
with a $50/50$\% probability.

\setlength{\fboxsep}{3pt}
\def\wpatch{\raisebox{0.5ex}{\fcolorbox{black}{white}{}}}
\def\gpatch{\raisebox{0.5ex}{\fcolorbox{black}{gray}{}}}
\def\bpatch{\raisebox{0.5ex}{\fcolorbox{black}{black}{}}}

\begin{table*}
  \caption{Quantitative overpainting comparisons over a test set of
    $180$ manually curated overpainting examples. (top) Mask adherence
    and editing quality measurements of prior methods trained on
    related tasks and applied to overpainting. (middle) Varying the
    attention-dropout rate trades mask adherence for editing
    quality. (bottom) Ablation results: (a) mask adherence is
    adversely affected when using only joint attention to integrate
    cues from the different input streams, and (b) training our model
    without inpainting training data improves overpainting performance
    at the cost of losing the ability to inpaint.}
  \label{tab:overpainting}
  \vspace{-3mm}
    % \footnotesize
    \centering
    \begin{tabular}{rlr|ccccc|cccc}
    \multicolumn{3}{c}{} & \multicolumn{5}{c}{Mask Adherence} & \multicolumn{4}{c}{Edit Quality} \\
    \toprule
				  & & & \% Aligned & \multicolumn{3}{c}{\% Edited pixels} & $F_1$↑ & CLIP↑ & EditCLIP↑ & AlphaCLIP↑ & EditReward↑ \\
				  & & & \;\;images↑ & \wpatch & \gpatch & \bpatch   & & & & &      \\
\midrule
\multicolumn{3}{r|}{FLUX Fill} & \underline{97.8} & 88.4 & 64.0 & \textbf{0.2} & 0.930 & 0.224 & 0.170 & 0.170 & 0.291 \\
\multicolumn{3}{r|}{FLUX (Prompted)} & 70.0 & 70.3 & 59.6 & 17.8 & 0.485 & \textbf{0.246} & 0.200 & 0.180 & 0.759 \\
\multicolumn{3}{r|}{FLUX (Annotated)} & 76.1 & 58.2 & 48.0 & 15.0 & 0.432 & \underline{0.243} & 0.198 & 0.180 & 0.539 \\
\multicolumn{3}{r|}{Nano Banana (Prompted)} & 56.1 & 78.0 & 58.1 & 12.9 & 0.591 & 0.241 & 0.209 & 0.186 & \textbf{1.063} \\
\multicolumn{3}{r|}{Nano Banana (Masked)} & 86.1 & 72.4 & 47.9 & 6.3 & 0.653 & 0.239 & \underline{0.214} & 0.186 & \underline{0.932} \\
\multicolumn{3}{r|}{UltraEdit} & \underline{97.8} & 76.9 & 55.9 & 2.5 & 0.776 & 0.229 & 0.187 & 0.174 & 0.310 \\
\multicolumn{3}{r|}{BrushNet} & 19.4 & \textbf{98.3} & 90.2 & 0.7 & \underline{0.968} & 0.234 & 0.184 & 0.175 & -0.314 \\
\multicolumn{3}{r|}{BrushEdit} & 20.6 & 91.5 & 86.3 & 1.3 & 0.912 & 0.217 & 0.155 & 0.156 & -0.567 \\
\multicolumn{3}{r|}{Blended Latent Diff.} & 92.8 & 92.4 & 75.6 & 0.3 & 0.947 & 0.218 & 0.152 & 0.159 & -0.368 \\
\multicolumn{3}{r|}{SDEdit (FLUX)} & \textbf{100.0} & 97.3 & 74.0 & \underline{0.2} & \textbf{0.978} & 0.224 & 0.175 & 0.170 & 0.114 \\
\midrule
\multirow{6}{*}{\rotatebox{90}{Attn. Dropout}} & 100\% & \textbf{Teamwork} & \textbf{100.0} & 96.2 & 54.2 & 0.4 & 0.962 & 0.236 & 0.202 & 0.191 & 0.650 \\
 & 75\% &  & \textbf{100.0} & \underline{97.5} & 53.9 & 0.7 & 0.956 & 0.238 & 0.203 & 0.190 & 0.682 \\
 & 50\% & \textbf{Ours} & \textbf{100.0} & 93.0 & 48.7 & 1.5 & 0.900 & 0.240 & 0.211 & 0.193 & 0.790 \\
 & 25\% &  & \textbf{100.0} & 91.1 & \underline{49.5} & 1.3 & 0.898 & 0.240 & 0.206 & 0.192 & 0.759 \\
 & 10\% &  & \textbf{100.0} & 88.0 & 46.1 & 1.9 & 0.859 & 0.239 & 0.209 & 0.191 & 0.849 \\
 & 0\% & \textbf{Naive} & \textbf{100.0} & 93.1 & \textbf{49.7} & 2.2 & 0.874 & 0.243 & \textbf{0.215} & \underline{0.193} & 0.903 \\
\midrule
\multicolumn{3}{r|}{FLUX + LoRA} & \textbf{100.0} & 77.8 & 53.9 & 16.7 & 0.499 & 0.239 & 0.202 & 0.179 & 0.744 \\
\multicolumn{3}{r|}{\textbf{Ours} (No Inpainting)} & \textbf{100.0} & 94.2 & 46.9 & 1.3 & 0.915 & 0.241 & 0.210 & \textbf{0.194} & 0.811 \\
    \bottomrule
    \end{tabular}
\end{table*}

\section{Evaluation}
\label{sec:eval}

\paragraph{Implementation}
We leverage the pre-trained FLUX.1 Kontext model which already includes joint-attention between image-streams.  We augment this model by applying Teamwork-LoRAs to all $W_Q$, $W_K$, $W_V$, $W_O$ matrices, $FF_1$ and $FF_2$, and the linear layer within each modulation block in FLUX.1 Kontext (but not the scaled-dot-product-attention itself). We use rank 64 Teamwork-LoRAs for all experiments.
To train our overpainting model , we use the standard diffusion loss
and the paired training images with corresponding prompts and
on-the-fly generated trimaps.  We train our model at 1024px resolution for $8,\!000$ steps with an effective batch size of 8. Training requires $\sim\!\!100$ hours of H100 compute. We use the standard settings of the Prodigy optimizer~\cite{Mishchenko:2023:PEA}with an initial learning rate of $1.0$, with cosine decay.

\begin{figure*}
    \centering
    \includegraphics[width=\textwidth]{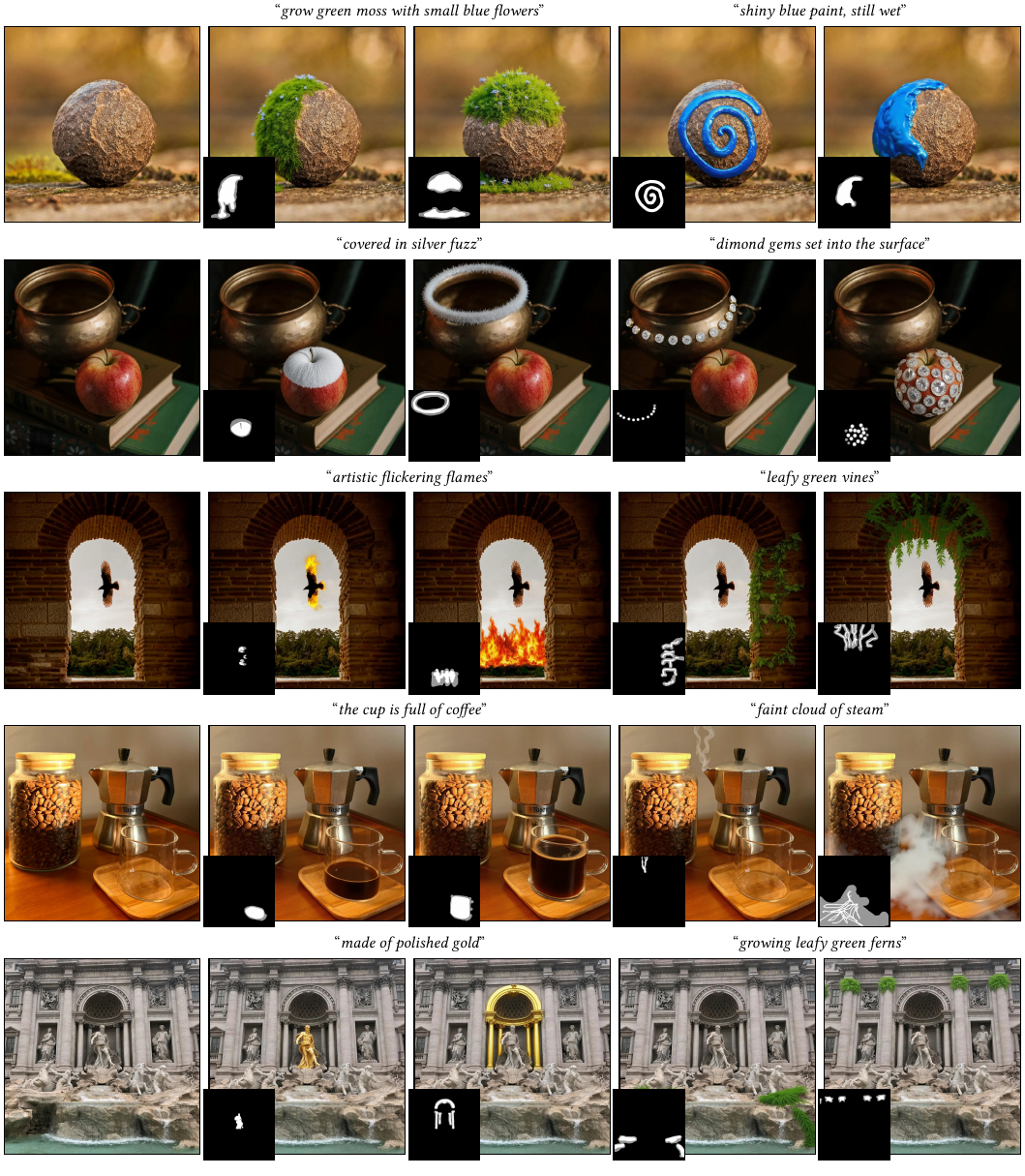}
    \caption{ Overpainting examples. For each input image, two
      different prompts (shown above the example), each with two
      different masks (shown as inset), are shown to demonstrate the
      rich variation in edits that are possible.  }
    \label{fig:results}%
\end{figure*}

\paragraph{Results}
\cref{fig:results} shows localized overpainting edits on a variety of
images, prompts, and masks. For each image, we apply four different
masks (inset) and two different prompts (shown above each example). We
include examples with very precise masks as well as examples with
masks with very large gray areas.  The support of different levels of
mask-accuracy allows users with varying skill levels to interact with
the model however they are most comfortable.  In some examples, the
masks are used to dictate the exact shape of the edit (\eg, spiral and
vines examples) whereas in other cases the mask is simply an
identifier of the region to edit and we leave it to overpainting to
find object boundaries (\eg, fire and smoke examples).  We also
observe that overpainting takes the context of the underlying object
in account when editing (\eg, grass and coffee examples).
Overpainting can perform \emph{material substitution} while respecting
the underlying shape (\eg, golden statue) as well as
\emph{alterations} where an effect is applied to an existing object or
scene (\eg, smoke example).  Overpainting is robust to sequential
editing (\cref{fig:teaser}) without significant degradation of the
unmasked regions.

\paragraph{Prior Work Comparison}
As overpainting is a new task, no competing model exists that solves
exactly the same task.  Instead, we repurpose existing models that
perform similar tasks to overpainting. We quantitatively compare our
overpainting model against FLUX Fill (an inpainting model), FLUX.1
Kontext and Nano Banana (image editing diffusion models) using either
(1) a more detail prompt describing the edit location (\eg, original:
\emph{"shiny blue tiles''}; detailed: \emph{"the area around her eyes
  is covered in shiny blue tiles''}), or (2) an annotated image with a
red outline or a mask with an appropriate prompt directing the model
how the use the added modality, and local diffusion editing models
(UltraEdit~\cite{Zhao:2024:UIB}, BrushNet~\cite{Ju:2024:BNP},
BrushEdit~\cite{Li:2024:BAI}, Blended
Diffusion~\cite{Avrahami:2023:BLD}), and SDEdit~\cite{Meng:2021:SDE}
with FLUX.1 Dev).

To evaluate the effectiveness of all models, we
manually author a diverse set of $180$ test cases. Each test case
consists of a starting image, hand-drawn mask, and an edit prompt.
In the test set, 5 cases have entirely black/gray masks with no white
at all, 15 cases have entirely white/black masks with no gray, 15 are
mostly gray with loosely placed white hints, 35 place gray strategically,
and 110 fully outline an otherwise coarse white selection. Test cases have
no ground-truth edit, so we quantitatively evaluate each method on
adherence to the input mask and on the quality of the resulting edit. 
See~\cref{fig:generalcmp} for a qualitative comparison on some examples.

%We also include the
%entire test set in supplemental and will release it as a benchmark
%upon publication \TODO{check this last statement is true}. \PP{We already mention in the intro that we will release everything}
For mask adherence, we measure alignment as the percentage of edited
images that are aligned with the source (using the method used during
training data generation), editing rate per mask-region as the total
percentage of pixels altered in each region (ideally this should be
$100\%, \sim\!\!50\%$, and $0\%$ for the white, gray, and black
regions, respectively), and the $F_1$ harmonic mean of precision and
recall, where the precision is the fraction of edited pixels which are
white or gray and the recall is the fraction of white pixels which are
edited.  For editing quality, we leverage existing metrics:
CLIP-IQA~\cite{Wang:2023:ECA}, EditCLIP~\cite{Wang:2025:ECR},
AlphaCLIP~\cite{Sun:2024:ACC}, and EditReward~\cite{Wu:2026:EHA} (a
VLM based metric).  From~\cref{tab:overpainting}, we observe that
while no single model excels in all metrics, our overpainting model
strikes a good balance between edit quality and mask adherence.

\begin{figure}
    \centering
    \includegraphics[width=\linewidth]{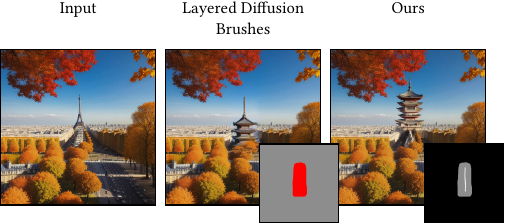}
    \vspace{-6mm}
    \caption{Compared against Layered Diffusion
      Brushes~\cite{Gholami:2025:SIE}, overpainting better retains the
      background and only edits the foreground object.  In this
      example, Layered Diffusion Brushes adds a blue glow around the
      tower and modifies the shape of the trees at the tower base.}
    \label{fig:ldbcmp}%
\end{figure}

\begin{figure}
    \centering
    \includegraphics[width=\linewidth]{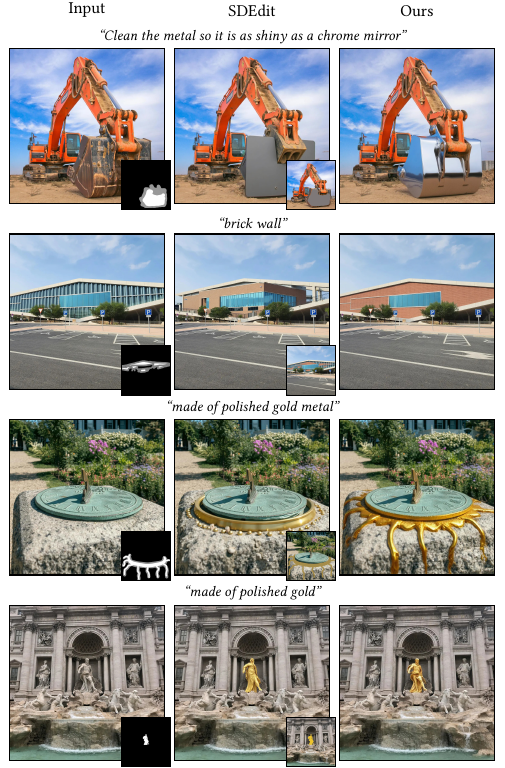}
    %\vspace{-6mm}
    \caption{A comparison against SDEdit~\cite{Meng:2021:SDE} using
      FLUX.1 Dev. We paint a solid color where the mask is white (shown in the inset)
      and denoise with 0.75 strength, while fixing the latents in
      black regions. We found that with lower strength the surface
      looks too flat, while higher strength further degrades identity
      preservation.}
    \label{fig:sdeditcmp}%
\end{figure}

Compared to other local editing methods, overpainting provides more
convenient ROI selection as well as better preservation of the context
of the edited region.  In~\cref{fig:ldbcmp}, we contrast the ease
of specifying a trimap in overpainting to Layered Diffusion
Brushes~\cite{Gholami:2025:SIE} where a very accurate mask is
required; without an accurate mask, the object might be incompletely
modified or the background can be affected.  \cref{fig:sdeditcmp}
demonstrates that SDEdit~\cite{Meng:2021:SDE} is capable of making
local edits similar to overpainting, but it requires careful fine-tuning
of hyper-parameters to balance identity-preservation and editing
strength.  Finally, Part\-Edit~\cite{Cvejic:2025:PFG}, due to its
mask-free interface, only supports a limited number of part-types and
it suffers from loss of precision in localizing edits
(\cref{fig:parteditcmp}).

\begin{table}
  \caption{Quantitative inpainting comparison of our model versus
    various inpainting models over the MISATO@1K
    dataset~\cite{Wang:2025:TEI} using LPIPS~\cite{Zhang:2018:TUE},
    FID~\cite{Heusel:2017:GTT}, and Paired/Unpaired Inception
    Discriminative Score (P-IDS/U-IDS)~\cite{Zhao:2021:LSI}.}
    \label{tab:inpainting}
    \footnotesize
    \centering
    \begin{tabular}{r|cccccc}
    \toprule
				      & U-IDS↑ & P-IDS↑ & LPIPS↓ & FID↓  &  Grad↓ & CLIP↑ \\
    \midrule
    FLUX-Fill         &  0.245  &  0.128  & \textul{0.165} & \textul{15.10}&   48.87 & \textbf{0.944} \\
    ASUKA (FLUX-Fill) & \textbf{0.311} & \textbf{0.185} & \textbf{0.149} & \textbf{13.50}&  \textbf{34.39}& \textul{0.942} \\
    ControlNet (SD3)  &  0.023  &  0.002  &  0.298  &  56.09 &   67.77 &  0.838 \\
    Teamwork (SD3)    &  0.116  &  0.041  &  0.204  &  25.20 &   48.47 &  0.900 \\
    FLUX-Kontext      &  0.025  &  0.013  &  0.372  &  92.09 &  241.41 &  0.833 \\
    \emph{Ours} (No Inpainting) & 0.000 & 0.000 &  0.475  & 151.56 &  183.02 &  0.722 \\
    \textbf{Ours}     & \textul{0.295} & \textul{0.131} &  0.189  &  20.58 &  \textul{42.39}&  0.923 \\
    \bottomrule
    \end{tabular}
\end{table}

\begin{figure*}
  \includegraphics[width=\textwidth]{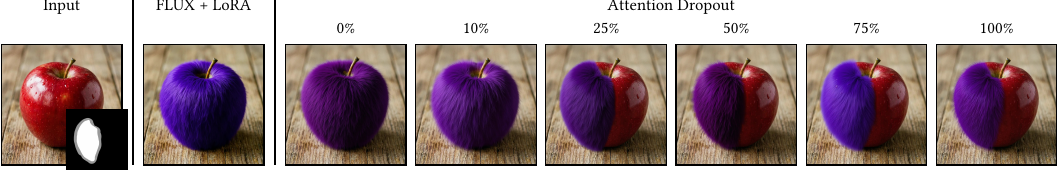}
  \vspace{-6mm}
  \caption{Varying the attention dropout rate trades-off mask
    adherence versus edit quality.  At low dropout rates, the mask is
    often ignored (\eg, $0\%$ and $10\%$). At higher dropout rates
    ($75\%$ and $100\%$), the editing quality degrades (\eg, the fur
    whiskers protruding from the silhouette appear less plausible). Our
    model ($50\%$ dropout) strikes a balance between mask adherence
    and editing quality.}
    \label{fig:dropout}
\end{figure*}

\begin{figure}
    \centering
    \includegraphics[width=\columnwidth]{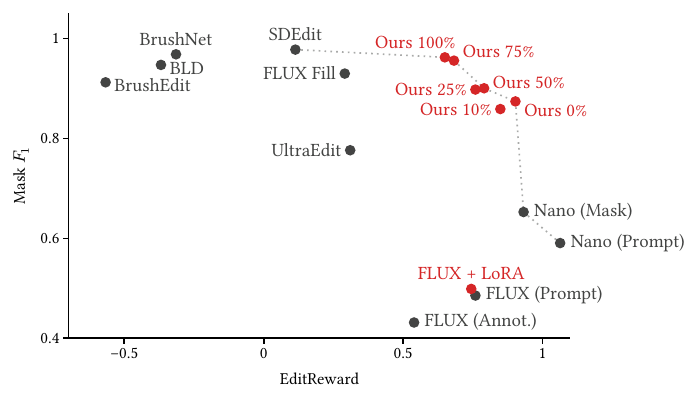}
    \vspace{-8mm}
    \caption{Pareto frontier (dotted line) of $F_1$ mask adherence
      versus EditReward across prior methods and our model with
      different attention-dropout rates.}
    \label{fig:paretofront}%
\end{figure}

\paragraph{Edit Quality vs. Mask Adherence}
As indicated in~\cref{sec:method}, there exists an inherent
trade-off between editing quality and mask adherence.  As
quantitatively shown in \cref{tab:overpainting}, FLUX+LoRA (\ie,
pure joint attention) achieves high marks on quality, but underperforms
in mask adherence, while for Teamwork ($100\%$ dropout) the opposite
is true. The naive combination ($0\%$ dropout) yields better quality
but at reduced mask adherence compared to the Teamwork-only model.
While the difference in mask adherence appears numerically minor,
it makes a significant difference qualitatively
(\cref{fig:dropout}), with an increased number of cases where
the mask is ignored.  Attention-dropout allow us to balance the
influence of joint attention and Teamwork. %
Increasing attention-dropout generally results in better mask
adherence at the cost of editing quality, as demonstrated
quantitatively in~\cref{tab:overpainting}. Plotting mask $F_1$
versus EditReward (\cref{fig:paretofront}), demonstrates that the
different dropout variants form a Pareto frontier and no single best
setting exists.  Empirically, we find that the $50\%$-dropout variant
strikes a good balance between mask adherence (without catastrophic
cases where the mask is ignored) and edit quality
(\cref{fig:dropout}).

\paragraph{Inpainting Comparison}
Overpainting and inpainting are distinct tasks.
An inpainting model expects the masked content to be erased, while an
overpainting model expects it to be present. However, by training on
a mix of both tasks, our model achieves comparable
inpainting quality to FLUX-Fill. \cref{fig:inpaintcmp} and \cref{tab:inpainting}
compare overpainting's inpainting capabilities qualitatively and
quantitatively to dedicated inpainting models. An interesting avenue for future research would be to apply ASUKA's~\cite{Wang:2025:TEI}
enhancement strategy to improve
inpainting quality.

For comparison, we also include overpainting (\cref{fig:inpaintcmp},
last column) to further illustrate the difference between overpainting
and inpainting: in the example of tiling the girl's face, we observe
that the inpainting models modify the identity of the subject. In
contrast, overpainting better retains the identity of the masked
region.

It should be noted that a specialized overpainting model
outperforms the combined inpainting/overpainting model in
quality (compare \cref{tab:overpainting} ``ours'' with/without
inpainting). If the best overpainting quality is desired,
we recommend to train our model without inpainting training
data, at the cost of loosing the ability to inpaint
(\cref{tab:inpainting} ``no inpainting''). All results in
this paper are generated with the inpainting-capable overpainting model.

\paragraph{Trimap Usability Study}
To validate the effectiveness of using a trimap as a masking
interface for directing overpainting, we perform a usability study in
which $11$ participants were asked to replicate a given target image
by drawing a mask given the input image, fixed prompt and fixed seed.
Each participant was tasked to replicate the target once with a binary
mask and once with a trimap (in random order). To keep the study brief
($< 15$min.), we provided only $4$ images; the result for the first
image was discarded as training example for the participant to get
hands-on experience with the system (a tutorial was also provided at
the beginning of the experiment). After each example, the participant
was asked to indicate their preference for either the binary or trimap
workflow on a 5-point scale.  Out of $11$ participants, $7$
participants preferred the trimap workflow (averaged over the $3$ test
cases), $1$ was neutral, and $3$ preferred the binary mask workflow.
Furthermore, while the degree of preference varies per test case, we
observe that support for the trimap workflow is strongest when target
region has intricate details or complex occlusions.  We refer to the
supplementary material for more details.

\begin{figure}
    \centering
\includegraphics[width=\linewidth]{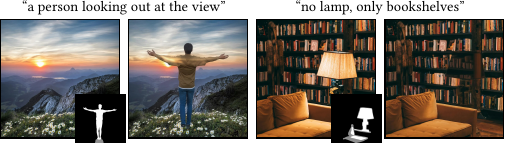}
\vspace{-6mm}
\captionof{figure}{Limitations: Overpainting is not suited for tasks
  where the context of the masked region is less important, such as
  when adding human subjects or removing objects.}
    \label{fig:limitations}%
\end{figure}

\paragraph{Limitations}
Our overpainting model is not without limitations.  First, we observe
that there is subjectively some quality loss compared to the base
model due to constraining the changes to a restricted region. Second,
our current overpainting model is sometimes unsuccessful when removing an
object or when adding a human subject (\cref{fig:limitations}).
However, neither are strictly overpainting tasks as in both cases the
context of the background does not matter.  Moreover, in the case of
human subjects the shape of the mask may impose too strong constraints
on the shape and pose of the subject and thus fall outside the learned
space.

%%%%%%%%%%%%%%%%%%%%%%%%%%%%%%%%%%%%%%%%%%%%%%%%%%%%%%%%%%%%
\section{Conclusion}
%%%%%%%%%%%%%%%%%%%%%%%%%%%%%%%%%%%%%%%%%%%%%%%%%%%%%%%%%%%%

In this paper we introduced ``overpainting'', a novel diffusion-based
method for making prompt-driven localized edits in an image while
taking the context of the edited region into account. Overpainting
differs from prior work on localized editing in its specification of
the region of interest via a trimap, allowing the user to trade-off
accuracy of the region of interest versus the precision of the edit
localization.  We showed that there exists an inherent trade-off
between mask adherence and edit quality, and we balance the
trade-off in our combined Teamwork - joint attention model via
attention-dropout.  We described an automated training data generation
pipeline that: (1)~generates a set of candidate image pairs leveraging
a pretrained diffusion editing model, (2)~automatically curates the
generated image pairs to retain only high quality training pairs, and
(3)~robustly extracts a mask/trimap from the image pairs.  Finally, we
demonstrated the efficacy of our overpainting model on a variety of
editing operations that are difficult to attain with prior editing
models.

%%%%%%%%%%%%%%%%%%%%%%%%%%%%%%%%%%%%%%%%%%%%%%%%%%%%%%%%%%%%
% References
%%%%%%%%%%%%%%%%%%%%%%%%%%%%%%%%%%%%%%%%%%%%%%%%%%%%%%%%%%%%

\bibliographystyle{ACM-Reference-Format}
\bibliography{references}
~\\

\clearpage

\clearpage
\onecolumn

\begin{minipage}{\textwidth}
%\begin{figure*}
    \centering
    \includegraphics{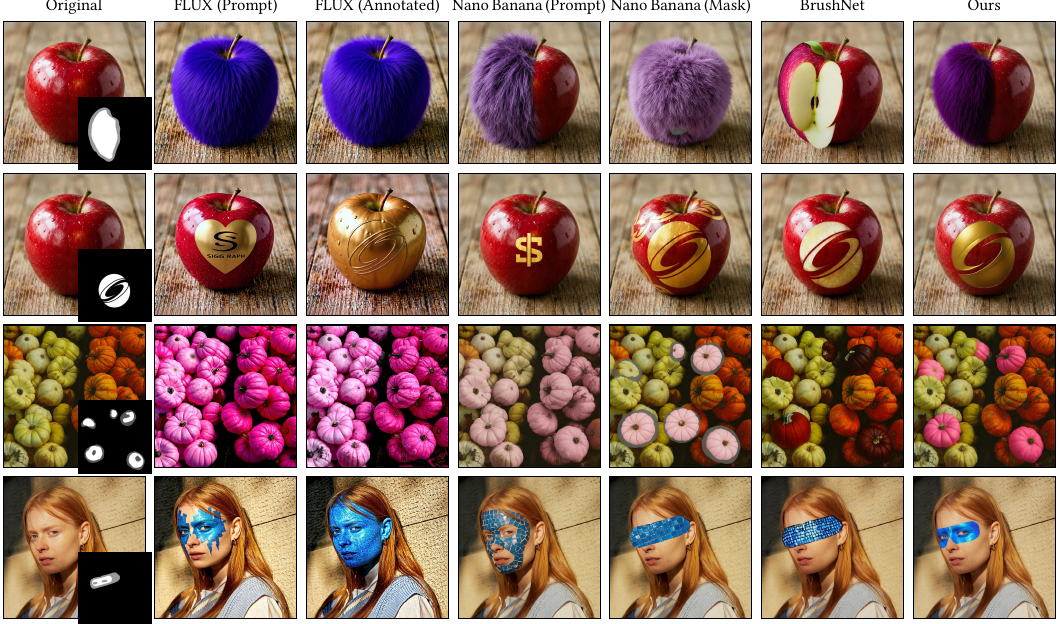}
    \vspace{-6mm}
    \captionof{figure}{Examples for overpainting with FLUX.1 Kontext using a
      detailed prompt (1st column), FLUX.1 Kontext with red outline
      annotations on the input image to demarcate the mask boundary
      (2nd column), Nano Banana with a detailed prompt (3rd column),
      Nano Banana with passing the mask as an additional input (4th
      column), BrushNet~\cite{Ju:2024:BNP} (5th column), and
      overpainting (6th column). Only overpainting adheres to the mask
      and the prompt in all cases.}
    \label{fig:generalcmp}%
%\end{figure*}
 
\end{minipage}

\vspace{1mm}
\begin{minipage}{0.47\textwidth}
  %\begin{figure}
    \centering
    \includegraphics[width=\linewidth]{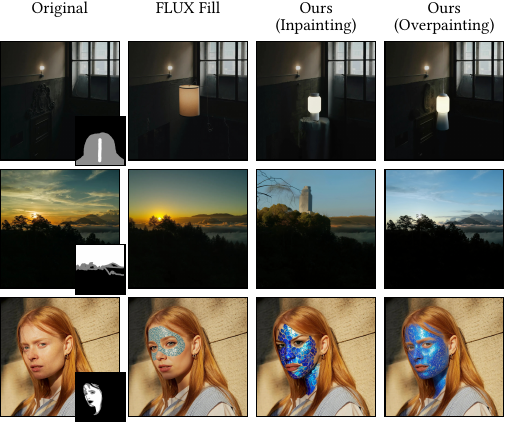}
    \vspace{-6mm}
    \captionof{figure}{Inpainting comparison of FLUX
      Fill against our model using the 'inpaint' keywords and
      overpainting.  Our model's inpainting results are of comparable
      quality to FLUX Fill.  However, both inpainting models fail to
      preserve the identity of the target region (\eg, the molding on
      the wall is lost (1st row), the mountains are lost or changed
      in the background when changing the sky to clear (2nd row), and
      the identity of the girl (3rd row) is altered). In contrast,
      overpainting preserves the identity of the target.}
    \label{fig:inpaintcmp}%
%\end{figure}

\end{minipage}
\hfill  
\begin{minipage}{0.47\textwidth}
  %\begin{figure}
    %\centering
    \includegraphics[width=\linewidth]{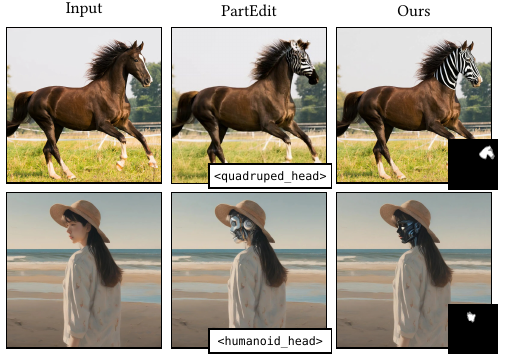}
    \captionof{figure}{Comparison between PartEdit~\cite{Cvejic:2025:PFG} and
      overpainting on editing a head (one of the recognized parts
      supported by PartEdit). While PartEdit can identify and edit the
      part, it fails to preserve the identity of the subject in
      contrast to overpainting.}
    \label{fig:parteditcmp}%
%\end{figure}

\end{minipage}

\pagebreak

\appendix
\twocolumn

\section{Architecture Details}

\begin{figure}
    \centering
    \includegraphics[width=0.8\linewidth]{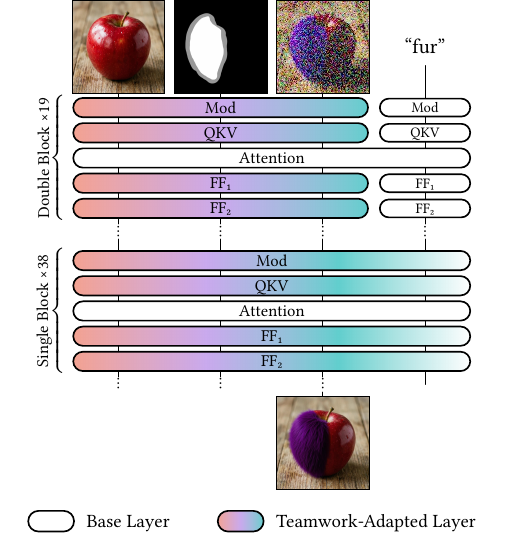}
    \caption{Architecture overview. We pass the source image
    and trimap as additional condition images, and communicate
    across the conditions and the noise using both teamwork and
    joint attention.
	}
    \label{fig:arch}%
\end{figure}

We demonstrate overpainting by finetuning the FLUX Kontext \cite{BFL:2025:FKF} model. FLUX Kontext
accepts latents for any number of condition images, as well as noisy latents for the image to generate,
all concatenated along the sequence dimension.

As a diffusion transformer (DiT) \cite{Peebles:2023:SDM}, FLUX Kontext consists of a series of 57 transformer
blocks which process all image patches and text tokens as a single sequence. Every transformer block contains:
\begin{enumerate}
    \item A modulation layer which produces scale and shift coefficients for following AdaLN normalization layers.
    \item An attention block using traditional multi-head scaled dot-product attention, and the usual linear $W_Q$, $W_K$, $W_V$, and $W_O$ projections.
    \item A feed-forward block which applies an MLP to every image/text feature, having two linear layers and a GeLU activation between them.
\end{enumerate}

The first 19 blocks of FLUX Kontext are ``Double Blocks'', where text tokens and image patches are separate streams,
each with their own layers, sharing only the scaled dot-product attention. The remaining 38 blocks are ``Single Blocks''
which use the same parameters for both the image stream and the text stream, treating both as a single uniform sequence.

There is some difficulty in applying Teamwork to FLUX Kontext, since the ``Batch Trick`` introduced by 
\citet{Sartor:2025:TCD} requires coordinated images to be arranged along the batch dimension.
To accommodate the batch trick, we adjust the Diffusers \cite{Platen:2022:DSD} implementation of
FLUX Kontext to accept image patches as a batch-wise $3b \times hw/p^2 \times n$ tensor, reshape to
a sequence-wise $b \times 3hw/p^2 \times n$ tensor prior to each scaled dot-product attention operation,
and reshape back to batch-wise afterwards. We also altered the single block implementation to compute
text and image features separately (using shared parameters), without Teamwork aggregation on text tokens.
Alternatively, our Teamwork model could be implemented without the batch-trick, keeping the sequence-wise shape
everywhere and grouping image patches by spatial location only within the Teamwork LoRAs. Either way, a
random block matrix is suitable as an attention mask to apply attention dropout during training.

Both our Teamwork and FLUX + LoRA overpainting finetunes apply LoRAs to all linear layers of the FLUX Kontext model, excluding only the input/ouput projections, timestep+guidance embedding projection, and text stream layers within
the 19 Double Blocks.

FLUX Kontext uses 3D RoPE embeddings \cite{Su:2024:ETR}
to disambiguate which image each latent patch belongs to and the spatial position of the patch within that
image. Base FLUX Kontext uses image index $0$ for the noisy latents and $1$ for the source image to edit.
We reuse index $1$ for our source image and add index $2$ for the mask image. Note the mask image is
VAE encoded in the same manner as any other condition image.

\section{Automated Training Data Generation -- Extended Description}

\paragraph{Paired Image Generation}
To train our overpainting model we require paired data in the form of
an original image, an edited version, a mask, and a prompt describing
the edit operation. Manually creating such training data at scale is
difficult., Therefore, we follow the well-established process of
leveraging VLMs and image diffusion models to generate training
data~\cite{Brooks:2023:ILF,
  Zhang:2023:MMA,Wasserman:2025:PBI,Qian:2025:PB4}. In our
implementation the training-data generation model is the same as the
overpainting base model (\ie, FLUX.1 Kontext), however, there is
nothing in our pipeline that requires that both need to be the same.

We start by randomly sampling $11,\!323$ freely licensed photographs
from \url{www.pexels.com} as source photographs. Each image is then
passed to the Qwen2.5-VL-72B multimodal language
model~\cite{Wu:2025:QIT} which is asked to suggest possible editing
prompts.  To promote diversity, we first seed Qwen's chain-of-thought
context by asking it to brainstorm twenty random 1- or 2-word concepts
that include objects, materials, colors, style, or themes.  Once
Qwen's context is seeded, we ask it to generate five meaningful
editing prompts of a few sentences, followed by a request to
abbreviate the prompt to keep it concise (\autoref{fig:prompt}).

Next, we feed the generated prompts and the corresponding source image
into FLUX.1 Kontext to produce the edited photographs.  To further
improve data quality and diversity, we also hand-select $60$ source
images and manually write editing prompts.  We leave the manually
written prompt intentionally ambiguous in order to run the same
image-prompt pair multiple times through FLUX.1 Kontext with different
seeds to generate diverse positional variations.  In total, we
generate $22,\!822$ automatically generated pairs and $1,\!000$ manual
pairs.

\paragraph{Filtering}
While powerful, FLUX.1 Kontext is not perfectly reliable and its
outputs exhibit several common failure modes.  We found that on
average $6.4\%$ of the generated results are nearly identical to the
original image (\ie, under-editing). In $7.1\%$ of the cases, FLUX.1
Kontext ignores the source image and generates an entirely new image
or makes drastic stylistic global changes without positional locality
(\ie, over-editing).  While the remainder of the results feature a
suitable edit, $42.7\%$ exhibit an additional global non-affine
shift of the whole image.  Such misalignment is not specific to FLUX.1
Kontext; we found that other models such as Google's Nano Banana
exhibit a similar misalignment.

We filter out over and under-edited results by finding the most and
least changed $16 \times 16$ image patch and measure the factor of
pixels with more than 0.15 change; for the most-changed patch we
require a minimum $75\%$ ratio and for the least-changed less than
$15\%$ should be changed.  This ensures that the edited image contains
regions with significant change and regions with minimal change.

To filter misaligned images, we extract matching key points using
LightGlue~\cite{Lindenberger:2023:LGL}, and reject pairs where more
than $15\%$ of the key points have moved more than $4$ pixels or if
too few ($<5$) matching key points are found.  We found that LightGlue
is able to reliably find matching features even in the presence of
moderate stylistic changes.  After filtering for misalignment we are
left with $43.8\%$ or $9,\!999$ good training pairs.

\begin{figure}
    \centering
    \includegraphics[width=\linewidth]{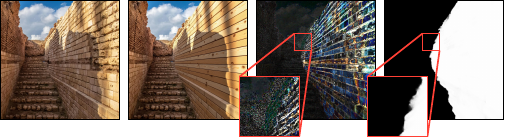}
    \caption{Example of degradation.  From left to right: the original
      image, the FLUX.1 Kontext edited image, the absolute pixel-wise
      difference, and the estimated mask.  As shown in the inset,
      FLUX.1 Kontext introduces small changes (\ie, degradation)
      outside the editing mask.}
      %%And because the overall color is similar before and after, edited areas do not necessarily have a higher pixel difference everywhere.   %%%% DID NOT INCLUDE THIS AS IT IS NOT THE MAIN MESSAGE OF THE IMAGE.
    \label{fig:degradation}%
\end{figure}

\begin{table}
  \caption{Quantitative ablation of mask estimation techniques. From right to left,
  the L1 difference of the estimated and ground-truth masks, the F1 similarity, and the
  intersection-over-union.}
    \label{tab:maskestimation}
    \footnotesize
    \centering
    \begin{tabular}{r|ccc}
    \toprule
                       &  L1↓  &  F1↑  &  I/U↑ \\
    \midrule
Thresholded Pixel Diff & 0.060 & 0.733 & 0.607 \\
  + Spatial Filter     & 0.042 & 0.841 & 0.758 \\
BiRefNet on Pixel Diff & 0.207 & 0.616 & 0.518 \\
  + Finetuning (ours)  & \textbf{0.007} & \textbf{0.955} & \textbf{0.916} \\
    \bottomrule
    \end{tabular}
\end{table}

\begin{figure}
    \centering
    \includegraphics[width=\linewidth]{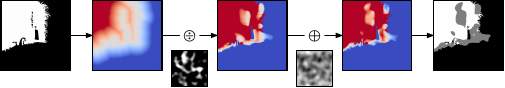}
    \caption{Given an alpha mask, we generate a trimap with varying
      gray regions by first estimating a signed distance field that
      we subsequently scale by a random thickness threshold (sampled
      from low-frequency noise), then offset by adding (another)
      low-frequency noise image (clamped positive), and finally threshold
      to three levels corresponding to the trimap regions.}
    % Our pipeline for augmenting estimated masks into imprecise
    % trimaps for training. We start by computing the signed distance
    % field of the estimated mask, sample random low-frequency noise
    % (clamped positive) to be the thickness of the ``may edit''
    % region, sample low-frequency noise to act as an offset within
    % that thickness, and then threshold to produce the black, white,
    % and grey mask. \TODO{There are two algebraicly equivalent ways
    % of doing this. (1) subtract the offset from the -1/1 thresholds,
    % scale by the thickness, and finally compare against the SDF. (2)
    % divide the SDF by the thickness, add the offset, threshold
    % against 1/-1. It is a lot easier to depict the later in a
    % figure, but I think the former makes more conceptual sense so I
    % kinda split the difference in this version of the caption.}}
    \label{fig:trimap}%
\end{figure}

\begin{figure}[t]
  \includegraphics[width=\linewidth]{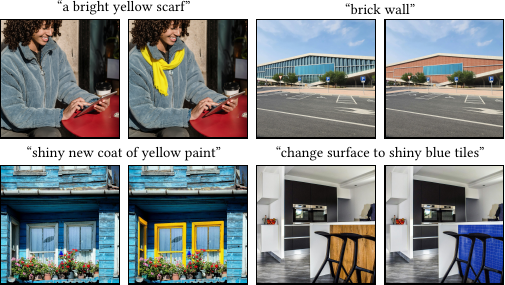}
  \caption{The four examples used in the usability study. For each example,
    we show the input image over which the participant draws the mask, and the
    target image shown to the participant as their goal. Each target image was
    made with the same overpainting model accessed by the participants, and with
    the same seed. The yellow scarf is the warm-up example for which the results
    are discarded.}
  \label{fig:studyimg}
\end{figure}

\paragraph{Mask Estimation}
While the paired images would be sufficient to train a global editing
model, overpainting requires an additional mask as input to indicate
where edits should occur. Since the above synthesis pipeline does not
generate such masks, we generate them post-hoc. The goal of ``mask estimation''
is to make this post-hoc binary prediction for each pixel: whether it was edited or not.

Naive thresholding is unreliable since image editing models tend
to introduce small, unintended differences to regions that are meant
to remain untouched; we refer to these undesired changes as
\emph{degradation} (\autoref{fig:degradation}).  Moreover, the edited
content can sometimes match the original color quite closely, even if the
texture is changed.  We therefore opt for a more robust mask
detection scheme by re-tasking BiRefNet~\cite{Zheng:2024:BRH} which
was originally trained for the related task of background removal, to
the task of mask detection.  We modify BiRefNet to take both the original
and edited image as input using head expansion.  We initialize the
weights of the expanded head as $[W, -W]$ (with $W$ the weights of
the original head), thereby starting the model from a state
similar to computing the pixel-wise difference between both inputs.

To train the mask estmimation model, we generate paired training data,
by randomly selecting an image from Pexels and extracting a foreground
and mask using the original (untuned) BiRefNet. Next, we composite the
extracted foreground on another randomly sampled (background) image
from Pexels. The composite, the original background, and composite
mask then serve as the training exemplars. To make the
mask estimation robust to degradation from the editing model, we apply
random brightness shifts and noise in both the background and
composited images in the FLUX VAE latent space.  We finetune the mask
estimation network for only $250$ steps. Due to the strong
BiRefNet-priors and the $[W, -W]$ head-initialization, the
mask-estimation network generalizes well to other edits.

In \autoref{tab:maskestimation} we provide a quantitative ablation of the
mask estimation model by computing the accuracy of 4 mask estimation
methods on 10 images edited by FLUX Kontext, against masks annotated
carefully by hand. The thresholded pixel diff proves unreliable,
even with spatial filtering to remove noise-induced holes/islands.
Immediately after head-expansion, BiRefNet does sometimes
identify the edited area simply because the large differences there look
more like "foreground" than the minimal noise elsewhere, but this is
also unreliable. A little finetuning on synthetic pairs (as described above)
obtains the best accuracy.

\autoref{fig:training} shows a selection of extracted binary masks and
corresponding image pairs.

\paragraph{Trimap generation}
The above mask-estimation network produces very precise alpha masks,
not trimaps.  While this accuracy is useful for exactly describing
where the edit must happen, we prefer less perfect masks to mimic
user-drawn trimaps.  To simulate this imprecision in mask
specification, we introduce randomized gray ``may edit'' regions
on-the-fly during training.  Intuitively, we extend the gray region
out by two different randomly selected distances on either side of the
boundary.  Practically, in our implementation we approximate this
efficiently as:
\begin{equation}
  \big| \, \text{sdf}(x) / \text{thickness}(x) + \text{offset}(x) \, \big| < 1,
\end{equation}
where $\text{sdf}(x)$ is the signed distance field with respect to the
black/whi\-te boundary, $\text{offset}(x)$ is a low-frequency noise
pattern clamped positive (range: $[0,1]$), and $\text{thickness}(x)$
is a low-frequency noise pattern (range: $[-t_{max}, +t_{max}]$, with
$t_{max}$ the maximum gray band thickness); see~\autoref{fig:trimap}
for a visualization of this process.  Note, the low-frequency
perturbations are critical to avoid generating trimaps where the
actual boundary always lies in the middle of the gray area; the model
would quickly overfit to this situation.

\section{Usability Study Details}
\begin{figure*}[ht]
    \centering
    \includegraphics[width=\textwidth]{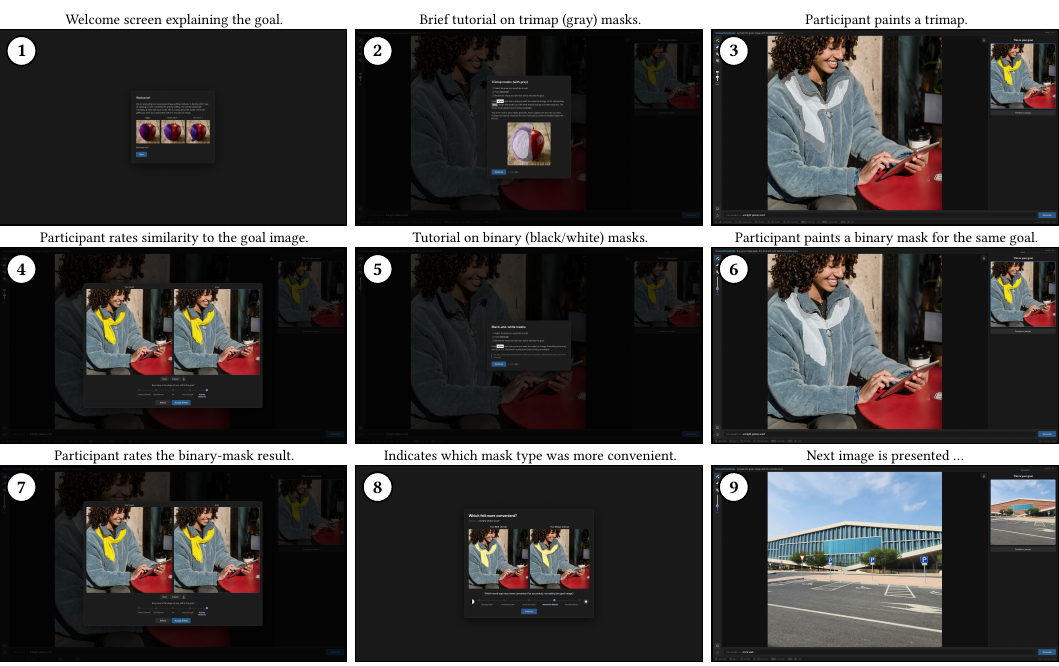}
    \caption{Walkthrough of one participant's session in our trimap
      usability study.}
    \label{fig:studyseq}%
\end{figure*}

\paragraph{Setup}
To validate the effectiveness of using a trimap as a masking
interface for directing overpainting, we perform a usability study in
which $11$ participants were asked to replicate a target image by
drawing a mask given the input image, fixed prompt and fixed seed.
Each participant was tasked to replicate the target once with a binary
mask and once with a trimap (in random order). To keep the study brief
($< 15$min.), we provided only $4$ images (\autoref{fig:studyimg});
the result for the first image was discarded as a warm-up example for
the participant to get hand-on experience with the system (a tutorial
was also provided at the beginning of the experiment).  We query the
participant to rate the result they obtained after each result in
order encourage the participant to produce high quality results.
After processing each example image with both mask interfaces, we ask
the participant to indicate their preference for either the binary or
trimap workflow on a 5-point scale.  \autoref{fig:studyseq} and the
supplementary video show a walkthrough of a participant's sesssion.

\begin{figure}[t]
  \includegraphics[width=0.7\linewidth]{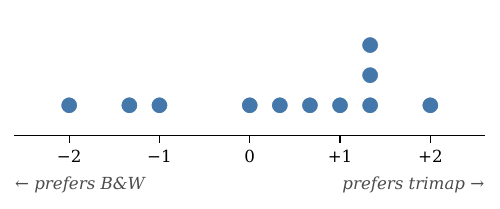}
  \caption{A Wilkinson plot of the per-participant average preference
    ratings, with a strong preference for the trimap on the right, and
    the binary mask on the left.}
  \label{fig:useravg}
\end{figure}

\begin{figure}
  \includegraphics[width=0.7\linewidth]{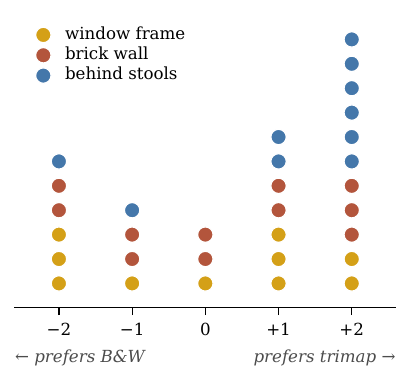}
  \caption{A Wilkinson plot of the preference ratings of participants
    for each example image, with a strong preference for the trimap on the
    right, and the binary mask on the left.}
  \label{fig:imgavg}
\end{figure}

\begin{figure}
  \includegraphics[width=0.7\linewidth]{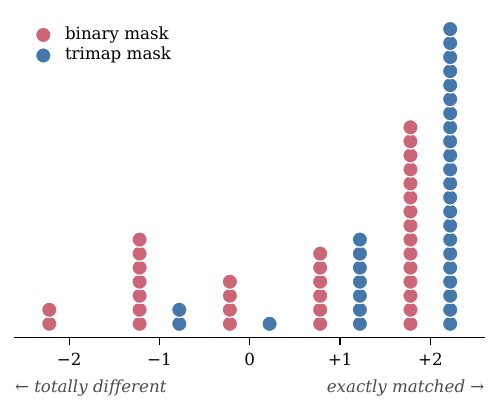}
  \caption{A Wilkinson plot of the similarity ratings of participants
    for trimaps versus binary masks, with higher self-rated similarity
    to the goal image on the right and lower on the left.}
  \label{fig:similaravg}
\end{figure}

\paragraph{Results Analysis}
\autoref{fig:useravg} shows a Wilkinson plot of the per-participant
average preference over the $3$ main examples used in the usability
study.  Only $3$ participants preferred the binary mask, while $7$
leaned more towards preferring the trimap.  However, we also observe
variability per participant between the different test
cases. \autoref{fig:imgavg} shows, for each example from the usability
study, a Wilkinson plot of the preference rating distribution.  As can
be seen, depending on the complexity of the image/mask, the preference
for using trimap varies.  However, in all cases, the trimap is the
most preferred interface.

During the usability study we also encouraged the participants to
produce high quality results via similarity self-rating. Plotting the
subjective similarity-rating per mask type (\autoref{fig:similaravg}),
reveals that the participants were able to obtain better matches with
trimaps than with binary masks.

\clearpage
% Two candidate layouts for the usability-study walkthrough.
% Comment out whichever variant you don't want once a choice is made.

\begin{figure*}
  \raggedright
  \footnotesize
  \textbf{System Prompt:}
  \begin{verbatim}
You are a creative brainstorming assistant.
Don't ever repeat yourself, instead come up with new ideas.
Avoid trivial, uninteresting, or very minor suggestions.
Be concise but unambiguous.
  \end{verbatim}

  \textbf{Seeding Prompt:}
  \begin{verbatim}
Hello!
I am creating a photo editing challenge for artists using photoshop.
Here is the photo in question.
To start brainstorming, can you give me {total_concepts} or so concepts you think would be interesting to explore?
Provide just one or two words for each.
Here are some random words as inspiration: {random_words}.
In your own list, include many different kinds of objects, materials, textures, colors, styles, and/or themes.
  \end{verbatim}

  \textbf{Prompt Generation Prompt:}
  \begin{verbatim}
Great!
Now can you propose {total_suggestions} edits contestants should make to this image?
Also let me know if the image has nudity, suggestive content, or violence (eg is nsfw).
Each challenge should require the artist to change the material, texture, or physical properties of one or more objects
in the scene. For example: changing a wooden table to a glass one, making a fabric look metallic, growing grass on dirt,
removing hair from a cat, covering a column with tree bark, etc.
Don't use these ideas directly, be inventive!
DO NOT ask for trivial color shifts, floating objects, or glowing effects.
Each change should be challenging to achieve and meaningfully alter the real physical contents of the scene.
Give specific information about what particular area of the image the artist should modify and what it should look like
after. But remain concise, instructions should be a few sentences each.
\end{verbatim}
  \caption{System, seeding, and prompt generation prompts provided to Qwen for generating possible overpainting editing operations.}
  \label{fig:prompt}
\end{figure*}

\end{document}